\documentclass[preprint,authoryear,12pt]{elsarticle}

\usepackage{amssymb}
\usepackage{array}
\usepackage{helvet}
\usepackage{courier}
\usepackage{url}
\usepackage{CJKutf8}
\usepackage{graphicx}
\usepackage{multirow}
\usepackage{amsmath}
\usepackage[normalem]{ulem}
\usepackage{amsopn}
\usepackage{times}
\usepackage{latexsym}
\usepackage[english]{babel}
\usepackage{bm}
\usepackage{float}
\usepackage{subfigure}
\usepackage{color}
\usepackage{algorithm}
\usepackage{algorithmic}
\usepackage{tikz}
\usepackage{tabularx}

\newcolumntype{C}[1]{>{\centering}p{#1}}



\journal{Artificial Intelligence}

\begin{document}

\begin{frontmatter}



\title{Enhanced Aspect-Based Sentiment Analysis Models with Progressive Self-supervised Attention Learning}


\author[mymainaddress]{Jinsong Su\corref{myequalcontribution}}
\ead{jssu@xmu.edu.cn}

\author[mysecondaddress,mythirdaddress]{Jialong Tang\corref{myequalcontribution}}
\ead{jialong2019g@iscas.ac.cn}

\author[mymainaddress]{Hui Jiang}
\ead{hjiang@stu.xmu.edu.cn}

\author[mymainaddress]{Ziyao Lu}
\ead{ziyaolu2018@stu.xmu.edu.cn}

\author[myfourthaddress]{Yubin Ge}
\ead{yubinge2@illinois.edu}

\author[mysixthaddress]{Linfeng Song}
\ead{lfsong@tencent.com}

\author[myfifthaddress]{\\Deyi Xiong\corref{mycorrespondingauthor}}
\ead{dyxiong@tju.edu.cn}

\author[mysecondaddress]{Le Sun}
\ead{sunle@iscas.ac.cn}

\author[mysixthaddress]{Jiebo Luo}
\ead{jluo@cs.rochester.edu}

\cortext[myequalcontribution]{Equal contribution}
\cortext[mycorrespondingauthor]{Corresponding author}

\address[mymainaddress]{Xiamen University, Xiamen 361005, China}
\address[mysecondaddress]{Institute of Software, Chinese Academy of Sciences, Beijing, China}
\address[mythirdaddress]{University of Chinese Academy of Sciences, Beijing, China}
\address[myfourthaddress]{University of Illinois at Urbana-Champaign, Urbana, IL 61801, USA}
\address[myfifthaddress]{Tianjin University, Tianjin, China}
\address[mysixthaddress]{University of Rochester, Rochester NY, USA}

\address{}

\begin{abstract}
In aspect-based sentiment analysis (ABSA),
many neural models are equipped with an attention mechanism to quantify the contribution of each context word to sentiment prediction.
However,
such a mechanism suffers from one drawback:
a few frequent words with sentiment polarities are tended to be considered while abundant infrequent sentiment words are ignored by the models.
To deal with this issue,
we propose a progressive self-supervised attention learning approach for attentional ABSA models.
We iteratively perform sentiment prediction on all training instances,
and continually extract useful attention supervision information in the meantime.
Specifically,
at each iteration,
the context word with the greatest effect on the sentiment prediction,
which is identified based on its attention weights or gradient,
is extracted as a word with active/misleading influence on the correct/incorrect prediction of every instance.
Words extracted in this way are masked for subsequent iterations.
To exploit these extracted words for refining ABSA models,
we augment the conventional training objective with a regularization term that encourages the models to not only take full advantage of the extracted active context words but also decrease the weights of those misleading words.
We integrate the proposed approach into three state-of-the-art neural ABSA models. 
Experiment results and in-depth analyses show that our approach yields better attention results and significantly enhances the performance of the models.
We release the source code and trained models at https://github.com/DeepLearnXMU/PSSAttention.
\end{abstract}


\end{frontmatter}

\section{Introduction}
As an indispensable and fine-grained task in sentiment analysis,
aspect-based sentiment analysis aims at automatically predicting the sentiment polarity (e.g. positive, negative, neutral) of an input sentence at the aspect level.
To achieve this goal,
early studies mainly resort to discriminative classifiers with manual feature engineering,
for example,
SVM-based ABSA models \citep{Kiritchenko:SemEval2014,Wagner:SemEval2014}.
Recent years have witnessed the rise to dominance of neural network-based ABSA models \citep{Tang:EMNLP2016,Wang:EMNLP2016,Tang:COLING2016,Ma:IJCAI2017,Chen:EMNLP2017,Li:ACL2018,Wang:ACL2018,Hu:ACL2019}.
Compared with previous traditional models,
neural ABSA models exhibit better performance due to their ability in learning the aspect-related semantic representations of input sentences.
Typically,
these models are usually equipped with an attention mechanism to quantify the contribution of each context word to the aspect-based sentiment prediction.
It cannot be denied that attention mechanisms play crucial roles in these dominant ABSA models.

However,
the existing attention mechanisms of ABSA models suffer from a major drawback,
which is also seen in neural models of other NLP tasks.
Specifically,
NNs are easily affected by these two patterns:
``\emph{apparent patterns}'' tend to be overly learned
while ``\emph{inapparent patterns}'' are not sufficiently learned \citep{Li:ACL2018,Xu:CONLL2018,Lin:ICCV2017}.
``\emph{Apparent patterns}'' and ``\emph{inapparent patterns}'' are widely present in the training corpus of ABSA,
where ``\emph{apparent patterns}'' are the high-frequency words with strong sentiment polarities while ``\emph{inapparent patterns}'' are low-frequency sentiment-related words.
Consequently,
attentional ABSA models tend to overly focus on high-frequency words with strong sentiment polarities
and little attention is laid upon low-frequency words,
leading to the unsatisfactory performance of these models.

\newcommand*{\MinNumberb}{0.0}
\newcommand*{\MaxNumberb}{0.5}
\newcommand{\Appb}[2]{\pgfmathsetmacro{\PercentColorb}{100.0*(#2-\MinNumberb)/(\MaxNumberb-\MinNumberb)}\colorbox{red!\PercentColorb!white}{\strut #1}}
\begin{table*}[!t]
\centering
\small
\resizebox{\textwidth}{18mm}{
\begin{tabularx}{17cm}{|c|X|c|}
\hline
{\bf Type} & \multicolumn{1}{|c|}{\bf Sentence} & {\bf Ans./Pred.}\\
\hline
\hline
{\bf Train} & \Appb{The}{0.1}\Appb{\bf [screen]}{0}\Appb{is}{0.2}\Appb{huge}{0.4}\Appb{and}{0.05}\Appb{colorful}{0.2}\Appb{but}{0.1}\Appb{no}{0.1}\Appb{led}{0.1}\Appb{back}{0.1}\Appb{\#\#light}{0.05}\Appb{\#\#ing}{0.1} & {\bf Pos / ---}\\
\hline
{\bf Train} & \Appb{The}{0.1}\Appb{\bf [graphics}{0}\Appb{\bf card]}{0}\Appb{is}{0.25}\Appb{a}{0.05}\Appb{huge}{0.45}\Appb{plus}{0.2}& {\bf Pos / ---}\\
\hline
{\bf Train} & \Appb{The}{0.05}\Appb{huge}{0.35}\Appb{\bf [screen]}{0}\Appb{allows}{0.2}\Appb{you}{0.05}\Appb{to}{0.05}\Appb{enjoy}{0.1}\Appb{watching}{0.2}\Appb{movies}{0.05}\Appb{,}{0.05}\Appb{pictures}{0.05}\Appb{and}{0.05}\Appb{etc}{0.05}& {\bf Pos / ---}\\
\hline
\hline
{\bf Test} & \Appb{The}{0.05}\Appb{\bf [performance]}{0}\Appb{of}{0.1}\Appb{mac}{0.05}\Appb{mini}{0.05}\Appb{is}{0.2}\Appb{a}{0.05}\Appb{huge}{0.45}\Appb{disappointment}{0.3}\Appb{.}{0.05}& {\bf Neg / Pos}\\
\hline
{\bf Test} & \Appb{The}{0.05}\Appb{\bf [screen]}{0}\Appb{looks}{0.15}\Appb{colorful}{0.1}\Appb{,}{0.05}\Appb{keyboard}{0.1}\Appb{is}{0.4}\Appb{not}{0.3}\Appb{very}{0.05}\Appb{sensitive}{0.05}& {\bf Pos / Neg}\\
\hline
\end{tabularx}}
\caption{\label{Table_Example1}
Examples of attention visualization for five sentences,
where the first three are training instances and the last two are test ones.
The bracketed bolded words are target aspects.
Ans. / Pred. = ground-truth/predicted sentiment label.
Words are highlighted with different degrees according to attention weights.
}
\end{table*}

Examples in Table \ref{Table_Example1} illustrate such phenomena.
During model training,
since the context word ``\emph{huge}" always occurs in the sentences with positive sentiment,
the attention mechanism pays more attention to it and directly relates sentences containing this word with positive sentiment.
Meanwhile,
another informative context word ``\emph{colorful}" tends to be partially neglected although it also possesses positive sentiment.
As a result,
during testing neural ABSA model incorrectly predicts the sentiment of the last two test sentences:
in the first sentence,
the attention mechanism directly focuses on ``\emph{huge}"
though it is not related to the given aspect;
while,
in the second sentence,
the neural ABSA model fails to capture the positive sentiment implicated by ``\emph{colorful}".
Apparently,
this defect restricts the performance of neural ABSA models.
Intuitively, the most direct solution to this issue is introducing supervised attention learning,
which, however, requires abundant manual annotation and thus is time-consuming.
Therefore,
It is still challenging to train a high-performance attention mechanism for ABSA models.

In this paper,
we propose a novel progressive self-supervised attention learning approach to enhancing attentional ABSA models.
Using this approach,
we can automatically and incrementally extract supervision information from the training corpus to guide the training of attention mechanisms in ABSA models.
The basic intuition behind our approach stems from the following fact:
for a training instance,
if the currently trained model can correctly predict its sentiment,
then its most important context word should be continuously considered in the subsequent model training.
By contrast,
this context word ought to be ignored.
To achieve this goal,
the most direct method is to extract this most important context as the supervised attention for the subsequent model training.
However,
as previously mentioned,
the most important context word is often the one with strong sentiment polarity.
It usually occurs frequently in the training corpus and thus tends to be overly considered during model training.
This simultaneously leads to the insufficient learning of other context words,
especially low-frequency ones with sentiment polarities.
To deal with this issue,
intuitively,
one possible way is to shield the effect of the most important context word and then re-predict the sentiment of its sentence,
in this way,
more influential contextual words even with low-frequency are expected to be extracted as attention supervision information.

Based on the above-mentioned analysis,
we iteratively conduct sentiment predictions on all training instances,
where various attention supervision information can be extracted to guide the training of attention mechanisms.
Specifically,
for each predicted training instance,
we extract different attention supervision information according to its predicted results:
if it can be correctly predicted,
we extract its most important context word and keep the extracted word to be considered for subsequent prediction;
otherwise, the attention weight of this extracted context word is expected to be decreased.
Then,
we mask all extracted context words of each training instance so far and then repeat the above process to discover more supervision information for attention mechanism.
Finally,
we augment the standard training objective with a regularizer,
which enforces attention distribution over these extracted context words to be consistent with their expected distribution.

To summarize,
the main contributions of our work are summarized as follows:
\begin{itemize}
\item
Through in-depth analysis,
we disclose the existing drawback of the attention mechanism for ABSA.

\item
We propose a novel incremental method to automatically extract attention supervision information for neural ABSA models.
To the best of our knowledge,
our work is the first attempt to explore automatic attention supervision information mining for ABSA.

\item
We apply our approach to three state-of-the-art neural ABSA models:
Memory Network (\textbf{MN}) \citep{Tang:EMNLP2016,Wang:ACL2018},
Transformation Network (\textbf{TNet}) \citep{Li:ACL2018}
and
BERT-based (\textbf{BERTABSA}) \citep{Hu:ACL2019} models.
Experimental results on several benchmark datasets
demonstrate the effectiveness of our approach.
\end{itemize}

Please note that this work has been presented in our previous conference paper \citep{Tang:ACL2019}.
In this article,
we make the following significant extensions to our previous work:
\begin{itemize}
\item
We propose a word saliency metric based on the partial gradient to identify the most important context word for our approach.
Experimental results across datasets show our new metric can refine our approach.

\item
We extend our proposed approach to a BERT-based ABSA model \citep{Hu:ACL2019}.
Experimental results show that the enhanced BERT-based ABSA model can achieve competitive performance,
verifying the generality of our approach on different neural attentional ABSA models.

\item
We conduct more experiments to further investigate the effectiveness and generality of our approach.
Besides,
we provide more details of our model,
such as the numbers of model parameters, as well as the model running time.
\end{itemize}

\section{Related Work}
Sentiment analysis has been widely studied recently \citep{Cambria:IIS2016,Hussain:NC2018,Vilares:SSCI2018,Chaturvedi:PRL2019}.
Among various sentiment analysis subtasks,
ABSA is an indispensable subtask that aims at inferring the sentiment polarity of an input sentence with respect to a certain aspect.
With the rapid development of deep learning, neural networks have been widely used and shown to be effective on ABSA \citep{Schouten:TKDE2016,Do:ESA2019,Zhou:IEEEAccess2019}.
In this aspect,
the typical neural ABSA models include
MNs with attention mechanisms \citep{Tang:EMNLP2016,Wang:ACL2018},
LSTMs that are equipped an attention mechanism to explicitly capture the importance of each context word \citep{Wang:EMNLP2016,Ma:AAAI2018},
and
hybrid approaches leveraging lexicalized domain ontology and neural attention models \citep{Wallaart:ESWC2019,Meskele:SAC2019,Meskele:IPM2020}.
Besides,
many researchers have conducted deep studies of this task from different perspectives,
such as aspect embedding \citep{Peng:KBS2018}, jointly modeling with other tasks \citep{Majumder:IIS2019,Majumder:AAAI2019}.
Particularly,
very recently,
BERT-based ABSA models \citep{Hu:ACL2019} have become the focus of ABSA study and exhibited state-of-the-art performance.

Obviously,
attention mechanisms are important components playing crucial roles in the above-mentioned neural ABSA models,
especially MN and LSTM-style ABSA models.
Along this line,
many researchers mainly focused on how to refine neural ABSA models via more sophisticated attention mechanisms.
For example,
\citet{Chen:EMNLP2017} propose a multiple-attention mechanism to capture sentiment features separated by a long distance,
which makes the model more robust against irrelevant information.
\citet{Ma:IJCAI2017} develop an interactive attention network for ABSA,
where two attention networks are introduced to model the target and context interactively.
\citet{Liu:EACL2017} propose to explore multiple attentions for ABSA:
one obtained from the left context and
the other one acquired from the right context of a given aspect.
Meanwhile,
the transformation-based model has also been explored for ABSA \citep{Li:ACL2018},
where, however, the attention mechanism is replaced by CNN.
Very recently,
BERT-based ABSA models have become the new state-of-the-art methods for ABSA.

Different from the above-mentioned work,
in this work,
we mainly focus on introducing attention supervision to refine the attention mechanism between the sentence hidden states and aspect representation.
In recent years,
task-related attention mechanisms been actively studied in several NN-based NLP tasks, such as
event detection \citep{Liu:ACL2017},
machine translation \citep{Liu:COLING2016,Zhang:TPAMI2018},
and police killing detection \citep{Nguyen:COLING2018}.
To achieve this goal,
the most direct approach is to introduce attention supervision information to refine attention mechanisms.
However,
the acquisition of such attention supervision information is labor-intense.
Therefore,
we mainly resort to automatically mining supervision information for attention mechanisms of neural ABSA models.
Theoretically,
our approach is orthogonal to these models. 
We leave the adaptation of our approach into these models as future work.

It should be noted that our work is partially inspired by two recent models:
one is \citep{Wei:CVPR2017} proposed to progressively mine discriminative object regions using classification networks to address the weakly-supervised semantic segmentation problems,
and the other is \citep{Xu:CONLL2018}
where a dropout method integrating with global information is presented to encourage the model to mine inapparent features or patterns for text classification.
Nevertheless,
to the best of our knowledge,
our work is the first effort to explore automatic mining of attention supervision information for ABSA.

\section{Background}\label{Section_Background}
In this work,
we choose MN \citep{Tang:EMNLP2016,Wang:ACL2018}, TNet \citep{Li:ACL2018} and BERTABSA \citep{Hu:ACL2019} as our basic ABSA models,
all of which are dominant neural ABSA models exhibiting satisfying performance.
Before we introduce these models,
we introduce some formal definitions:
$x$= ($x_1,x_2,...,x_N$) is the input sentence,
$t$= 
($t_{1},t_{2},...,t_{T}$) is the given target aspect,
$y$, $y_p$$\in$$\{$Positive, Negative, Neutral$\}$
denote the ground-truth and the predicted sentiment, respectively.

\subsection{\textbf{M}emory \textbf{N}etwork (\textbf{MN})}
\begin{figure}[!t]
\centering
\includegraphics[width=0.45\textwidth]{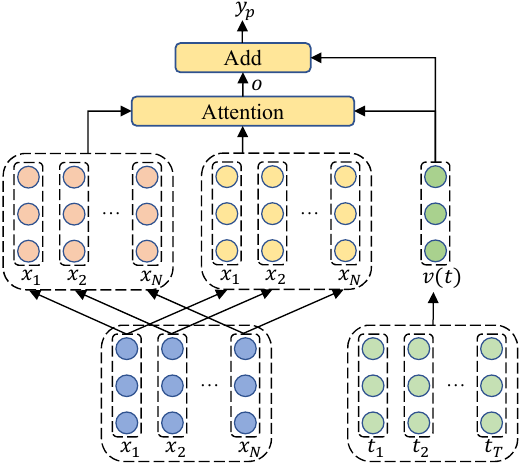}
\caption{\label{Fig_MN}
The architecture of MN.
}
\end{figure}

\begin{figure}[!t]
\centering
\subfigure[TNet]{
\label{Subfig_TNet}
\includegraphics[width=0.48\textwidth]{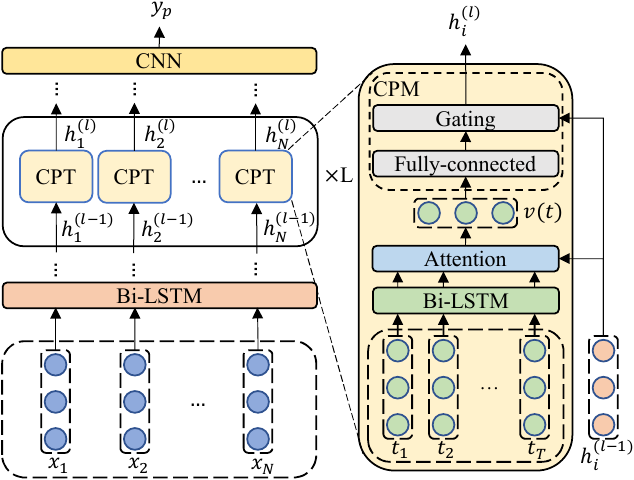}
}
\subfigure[TNet-ATT]{
\label{Subfig_TNetATT}
\includegraphics[width=0.48\textwidth]{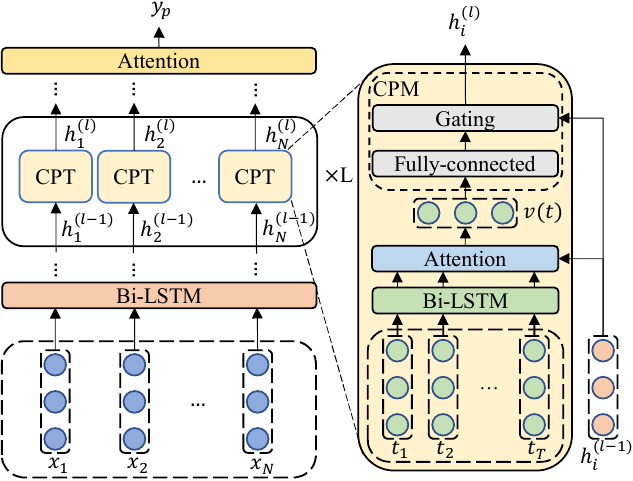}
}
\caption{The architectures of TNet and TNet-ATT.
Note that TNet-ATT is the variant of TNet, which replaces CNN with an attention mechanism to produce the aspect-related sentence representation.}
\label{Fig_TNets}
\end{figure}

Figure \ref{Fig_MN} shows the basic framework of MN.
In this model,
the vector representation $v(t)$ of target aspect $t$ is first defined as the averaged aspect embedding of its words,
all of which are indexed from an aspect embedding matrix.
Meanwhile,
we introduce a word embedding matrix projecting each context word $x_i$ to the continuous space stored in memory,
denoted by $m_i$.
Then,
an internal attention mechanism is applied to generate the aspect-related semantic representation $o$ of sentence $x$:
$o$ =$\sum^N_{i=1}$\emph{softmax}$(v(t)^\top M m_i) h_i$,
where $M$ is an attention matrix and $h_i$ is the final semantic vector representation projected from $x_i$ via another context word embedding matrix.
Finally,
we employ a fully connected output layer to conduct classification based on a summation vector over $o$ and $v(t)$.

\subsection{\textbf{T}ransformation \textbf{Net}work (\textbf{TNet})}
As shown in Figure \ref{Subfig_TNet},
the framework of TNet consists of three components: a bottom layer, a middle component and a CNN classification layer.
The bottom layer is a Bi-LSTM that transforms the input $x$ into the contextualized word representations $h^{(0)}(x)$=($h_1^{(0)},h_2^{(0)},...,h_N^{(0)}$) (i.e. hidden states of Bi-LSTM).
The middle component is the core of the whole model.
It consists of $L$ layers of Context-Preserving Transformation (CPT),
where word representations are updated as $h^{(l+1)}(x)$=CPT($h^{(l)}(x)$).
One key operation in the CPT layers is Target-Specific Transformation,
which contains another Bi-LSTM to generate $v(t)$ via an attention mechanism,
and then incorporates $v(t)$ into the word representations.
Besides,
each CPT layer is also equipped with a Context-Preserving Mechanism (CPM) to
preserve the context information and learn more abstract word-level features.
Finally,
we obtain the word-level semantic representations\\
$h(x)$$=$$h^{(L)}(x)$$=$($h^{(L)}_1(x)$,$h^{(L)}_2(x)$...,$h^{(L)}_N(x)$).
On the top of the CPT layers,
a CNN layer is used to produce the aspect-related sentence representation $o$ for the final sentiment classification.

Note that the original TNet uses CNN rather than attention mechanism to extract the sentence representation,
and thus our approach cannot be applied directly to the original TNet.
In this work,
we consider another alternative to the original TNet,
denoted as \textbf{TNet-ATT}.
As shown in \ref{Subfig_TNetATT},
it replaces its topmost CNN with an attention mechanism to produce the aspect-related sentence representation as $o$=\emph{Attention}($h(x)$,\ $v(t)$).
In the latter experiment section,
we will investigate the performance of the original TNet and TNet-ATT.

\subsection{\textbf{BERT}-based \textbf{ABSA} Model (\textbf{BERTABSA})} \label{SubSection_BERTbasedABSAModel}
\begin{figure}[!t]
\centering
\subfigure[BERTABSA]{
\label{Subfig_BERTABSA}
\includegraphics[width=0.48\textwidth]{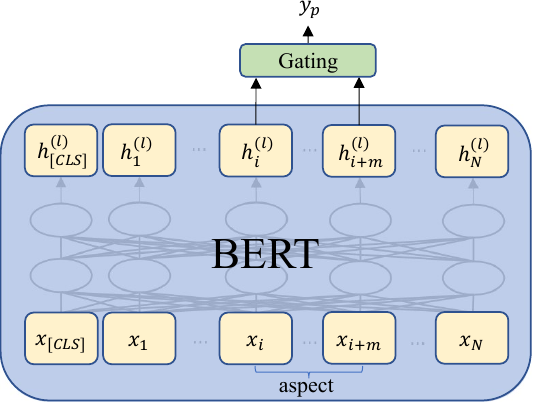}
}
\subfigure[BERTABSA-ATT]{
\label{Subfig_BERTABSAATT}
\includegraphics[width=0.48\textwidth]{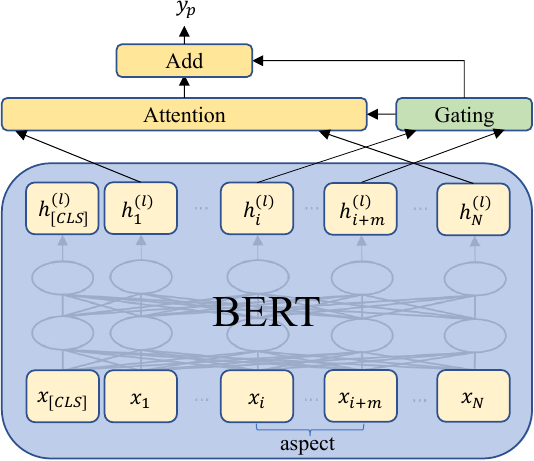}
}
\caption{The architectures of BERTABSA and BERTABSA-ATT.
BERTABSA-ATT is the variant of BERTABSA,
which is equipped with an attention mechanism to generate the aspect-related sentence representation.}
\label{Fig_BERTABSAs}
\end{figure}

The overall illustration of BERTABSA is shown in Figure \ref{Subfig_BERTABSA}.
Its basis is the BERT \citep{Devlin:NAACL2019} encoder,
which is a pre-trained bidirectional Transformer encoder and achieves state-of-the-art performance across a variety of NLP tasks.
Using this encoder,
we can produce the contextualized word representations $h(x)$ of input sentence $x$,
forming the basis of the final sentiment classification.

Specifically,
we first generate a new input sentence $\hat{x}$ by concatenating a [CLS] token, input sentence $x$ and a [SEP] token,
and represent each input word with a vector summing the word, segment and position embeddings.
Then,
we use $L$ layers of stacked Transformer blocks to project these input summed vectors into a sequence of contextualized word representations:
$h^{(l+1)}(x)$=\emph{Transformer}($h^{(l)}(x)$).
Here we omit the detailed description of the function \emph{Transformer}(*).
Please refer to \citep{Vaswani:NIPS2017} for more details.
Afterwards,
we employ a gating mechanism to produce the vector representation $v(t)$ of target aspect,
$v(t)$ =$\sum_{j=tb}^{te}$\emph{softmax}$(Wh^{(L)}_j)$$h^{(L)}_j$,
where $W$ is a gating matrix.
Finally,
we conduct the sentiment classification based on $v(t)$.

Likewise,
the original BERTABSA cannot be directly improved by our approach.
In this work,
following MN,
we also apply an attention mechanism to generate the aspect-related semantic representation $o$ =$\sum^N_{i=1}$\emph{softmax}$(\frac{v(t)^\top h_i}{\sqrt{d}}) h_i$,
as illustrated in Figure \ref{Subfig_BERTABSAATT}.
Here, $d$ indicates the dimension of the hidden state.
Finally,
we introduce a fully connected output layer to conduct classification based on the summing vector of $o$ and $v(t)$.

We refer to the original BERT-based ABSA model and its enhanced variant as \textbf{BERTABSA} and \textbf{BERTABSA-ATT}, respectively.
We also compare their performance in the later experiment section.

\subsection{Training Objective}
All above-mentioned models take the negative log-likelihood of the gold-truth sentiment tags as their training objectives.
Formally,
give the training corpus $D=(x,t,y;\theta)$,
the training objective is defined as follows:
\begin{align}\label{Eqa_OldTrainingObjective}
J(D;\theta) &= -\sum_{(x,t,y)\in D} J(x,t,y;\theta) \notag \\
            &=\sum_{(x,t,y)\in D} d(y) \cdot \textrm{log}d(x,t; \theta),
\end{align}where $d(y)$ is the one-hot vector of $y$,
$d(x,t;\theta)$ is the model-predicted sentiment distribution for the pair ($x$,$t$),
and ``$\cdot$'' denotes the dot product of two vectors.

{
\renewcommand
\baselinestretch{1.1}
\begin{algorithm}[!t]
\footnotesize
\renewcommand{\algorithmicrequire}{\textbf{Input:}}
\renewcommand\algorithmicensure {\textbf{Return:} }
\caption{: Neural ABSA Model Training with Automatically Mined Attention Supervision Information.}
\label{Alg_OurApproach}
\begin{algorithmic}[1]
\REQUIRE
$\textit{D}$: the initial training corpus;\\
$\theta^{init}$: the initial model parameters;\\
$\epsilon_{\alpha}$: the entropy threshold of attention weight distribution;\\
$\textit{K}$: the maximum number of training iterations;
\STATE $\theta^{(0)}$ $\leftarrow$ \textit{\textbf{Train}}($\textit{D}$;\ $\theta^{init}$)
\STATE \textbf{for} ($x$,\ $t$,\ $y$) $\in$ $\textit{D}$ \textbf{do}
\STATE \ \ \ \ \ \ $s_a(x)$ $\leftarrow$ $\emptyset$
\STATE \ \ \ \ \ \ $s_m(x)$ $\leftarrow$ $\emptyset$
\STATE \textbf{end for}
\STATE \textbf{for} $k=1,2...,\textit{K}$ \textbf{do}
\STATE \ \ \ \ \ \ $\textit{D}^{(k)}$ $\leftarrow$ $\emptyset$
\STATE \ \ \ \ \ \ \textbf{for} ($x$,\ $t$,\ $y$) $\in$ $\textit{D}$ \textbf{do}
\STATE \ \ \ \ \ \ \ \ \ \ \ \ $v(t)$ $\leftarrow$ $\textit{\textbf{GenAspectRep}}$($t$,\ $\theta^{(k-1)}$)
\STATE \ \ \ \ \ \ \ \ \ \ \ \ $x'$ $\leftarrow$ $\textit{\textbf{MaskWord}}$($x$,\ $s_a(x)$,\ $s_m(x)$)
\STATE \ \ \ \ \ \ \ \ \ \ \ \ $h(x')$  $\leftarrow$ $\textit{\textbf{GenWordRep}}$($x'$,\ $v(t)$,\ $\theta^{(k-1)}$)
\STATE \ \ \ \ \ \ \ \ \ \ \ \ $y_p$, $\alpha(x')$ $\leftarrow$ \textit{\textbf{SentiPred}}($h(x')$,\ $v(t)$,\ $\theta^{(k-1)}$)
\STATE \ \ \ \ \ \ \ \ \ \ \ \ $E(\alpha(x'))$ $\leftarrow$ \textit{\textbf{CalcEntropy}}($\alpha(x')$)
\STATE \ \ \ \ \ \ \ \ \ \ \ \ \textbf{if} $E(\alpha(x'))$ $<$ $\epsilon_{\alpha}$ \textbf{then}
\STATE \ \ \ \ \ \ \ \ \ \ \ \ \ \ \ \ \ \ \ $m$ $\leftarrow$ $argmax_{1\leq i \leq N}$ $\alpha(x'_i)$
\STATE \ \ \ \ \ \ \ \ \ \ \ \ \ \ \ \ \ \ \ \textbf{if} $y_p$ $==$ $y$ \textbf{then}
\STATE \ \ \ \ \ \ \ \ \ \ \ \ \ \ \ \ \ \ \ \ \ \ \ $s_a(x)$ $\leftarrow$ $s_a(x)$ $\cup$ $\{x'_m\}$
\STATE \ \ \ \ \ \ \ \ \ \ \ \ \ \ \ \ \ \ \ \textbf{else}
\STATE \ \ \ \ \ \ \ \ \ \ \ \ \ \ \ \ \ \ \ \ \ \ \ $s_m(x)$ $\leftarrow$ $s_m(x)$ $\cup$ $\{x'_m\}$
\STATE \ \ \ \ \ \ \ \ \ \ \ \ \ \ \ \ \ \ \ \textbf{end if}
\STATE \ \ \ \ \ \ \ \ \ \ \ \ \textbf{end if}
\STATE \ \ \ \ \ \ \ \ \ \ \ \ $\textit{D}^{(k)}$ $\leftarrow$ $\textit{D}^{(k)}$ $\cup$ ($x'$,\ $t$,\ $y$)
\STATE \ \ \ \ \ \ \textbf{end for}
\STATE \ \ \ \ \ \ $\theta^{(k)}$ $\leftarrow$ \textit{\textbf{Train}}($\textit{D}^{(k)}$;\ $\theta^{(k-1)}$)
\STATE \textbf{end for}
\STATE $\textit{D}_s$ $\leftarrow$ $\emptyset$
\STATE \textbf{for} ($x$,\ $t$,\ $y$) $\in$ $\textit{D}$ \textbf{do}
\STATE \ \ \ \ \ \ \ $\textit{D}_s$ $\leftarrow$ $\textit{D}_s$ $\cup$ ($x$,\ $t$,\ $y$,\ $s_a(x)$,\ $s_m(x)$)
\STATE \textbf{end for}
\STATE $\theta$ $\leftarrow$ \textit{\textbf{Train}}($\textit{D}_s$)
\ENSURE $\theta$;
\end{algorithmic}
\end{algorithm}
\par}

\section{Our Approach}\label{Section_OurApproach}
In this section,
we will elaborate our approach.
Please note that our method only affects the training optimization of neural ABSA models.
First,
we describe the procedure of automatically extracting attention supervision information from a training corpus.
Then,
we augment the conventional training objective with a regularizer to exploit this information.

\subsection{Extracting Attention Supervision Information}\label{SubSection_ExtractAttentionSupervision}

To facilitate the description of extracting attention supervision information,
we summarize this process in Algorithm \ref{Alg_OurApproach}.
We first use the initial training corpus $D$ to train the initial model with parameters $\theta^{(0)}$ (\textbf{Line 1}).
Then, we continue training the model for $K$ iterations,
where influential context words of all training instances can be iteratively extracted as attention supervision information (\textbf{Line 6-25}).
To iteratively extract attention supervision information,
for each training instance ($x,t,y$),
we introduce two word sets initialized as $\emptyset$ (\textbf{Lines 2-5}) to store its all extracted context words:
(1) $s_a(x)$ consists of context words with \textbf{a}ctive effects on the sentiment prediction of $x$.
Each word in $s_a(x)$ will be encouraged to remain considered in the refined model training,
and
(2) $s_m(x)$ contains context words with \textbf{m}isleading effects,
whose attention weights are expected to be decreased.
More specifically,
at the $k$-th training iteration,
we adopt the following steps to deal with ($x,t,y$):

In {\textbf{Step 1}},
we first use the model parameters $\theta^{(k-1)}$ of the previous iteration to generate
the aspect representation $v(t)$ (\textbf{Line 9}).
Then,
as a very important step,
according to $s_a(x)$ and $s_m(x)$,
we create a new sentence $x'$ by replacing each previously extracted word of $x$ with a special token ``$\langle mask\rangle$'' (\textbf{Line 10}).
In this way,
effects of these context words will be shielded
during the sentiment prediction of $x'$,
and thus other important context words can potentially be extracted from $x'$.
Finally,
we generate the word representations $h(x')$$=$$\{h(x'_i)\}^N_{i=1}$ (\textbf{Line 11}).

In \textbf{Step 2},
on the basis of $v(t)$ and $h(x')$,
we leverage $\theta^{(k-1)}$ to predict the sentiment of $x'$ as $y_p$ (\textbf{Line 12}).
During this process,
we can simultaneously induce a word saliency score vector
\footnote{In our conference paper \citep{Tang:ACL2019},
we directly define $\alpha(*)$ as an attention weight vector.
While,
in this paper,
we refer to $\alpha(*)$ as word saliency score vector which has two definitions.
}
$\alpha(x')$=$\{\alpha(x'_1),\alpha(x'_2),...,\alpha(x'_N)\}$
that is subject to $\sum^N_{i=1}\alpha{(x'_i)}=1$,
where $\alpha(x'_i)$ measures the influence of word $x'_i$ on the sentiment prediction of $x'$.
Here we explore two definitions to calculate $\alpha(x'_i)$,
which will be investigated in the latter experiment section.
\begin{itemize}
\item
\textbf{Definition 1}
Following \citep{Ghader:IJCNLP2017,Ghaeini:EMNLP2018},
we directly define $\alpha(x'_i)$ as the attention weight of $x'_i$,
which scores how well the semantic representation $h(x'_i)$ of $x'_i$ and the semantic representation $v(t)$ of the input aspect match.

\item
\textbf{Definition 2}
As analyzed in previous studies \citep{Jain:NAACL2019},
learned attention weights are frequently uncorrelated with gradient-based measures of feature importance.
To tackle this defect,
we follow \citep{Jain:NAACL2019, Serrano:ACL2019, Ding:WMT2019} to define $\alpha(x'_i)$ based on its partial gradient.

Specifically,
we first compute the partial gradients of the predicted sentiment distribution $d(x',t;\theta)$ with respect to each cell of $h(x'_i)$.
Then,
we calculate the weighted sum of $h(x'_i)$ with the partial gradient of each cell as the weight.
Finally,
we use the normalized absolute value of this weighted sum to define $\alpha(x'_i)$.

It should be noted that if the model fits the distribution perfectly,
some data points or input features may become saturated,
leading to a partial gradient of 0,
which, however, does not mean that they have no impact on the final prediction.
To deal with this issue,
we generate $n$ samples by adding random noise into the vector representation of $x_i$ according to the normal distribution $N(0, \sigma^2)$.
By doing so,
we can cancel out the noise in the gradients to obtain a more accurate saliency score by averaging the word-level saliency scores of all samples.
\end{itemize}

In \textbf{Step 3},
we first use the entropy $E(\alpha(x'))$ to measure the variance of $\alpha(x')$ (\textbf{Line 13}),
\begin{align}\label{Eqa_Entropy}
E(\alpha(x')) = -\sum^N_{i=1} \alpha(x'_i) \log(\alpha(x'_i)).
\end{align}Then, based on $E(\alpha(x'))$,
we can determine whether $x'$ contains any important context words contributing to the sentiment prediction of $x'$.
When $E(\alpha(x'))$ is less than a threshold $\epsilon_{\alpha}$ (\textbf{Line 14}),
we believe that there exists one such influential context word and extract the context word $x'_m$ with the maximum influence weight (\textbf{Line 15-20}) as attention supervision information.
To be specific,
we deal with $x'_m$ according to different prediction results on $x'$:
if the prediction is correct,
we wish to continue focusing on $x'_m$ and add it into $s_a(x)$ (\textbf{Lines 16-17});
otherwise,
we expect to decrease the attention weight of $x'_m$ and thus include it into $s_m(x)$ (\textbf{Lines 18-19}).

In \textbf{Step 4},
we combine $x'$, $t$ and $y$ as a triple,
and merge it with the collected ones to form a new training corpus $D^{(k)}$ (\textbf{Line 22}).
Then,
we leverage $D^{(k)}$ to continue updating model parameters for the next iteration (\textbf{Line 24}).
In doing so,
we make our model adaptive to discover more influential context words.

Through $K$ iterations of the above steps,
we will extract influential context words of all training instances.
Table \ref{Table_Example2a} and \ref{Table_Example2b} illustrate the procedures of extracting influential context words from the first sentence shown in Table \ref{Table_Example1}.
When using attention weights to define word-level saliency scores,
we can iteratively extract five context words in turn:
``\emph{huge}'', ``\emph{is}'', ``\emph{colorful}'', ``\emph{but}'', and ``\emph{no}''.
The former three words are included in $s_a(x)$,
while the last two are contained in $s_m(x)$.
If we use partial gradients to define word-level saliency scores,
the extracted context words also comprise two categories of words:
``\emph{huge}'', ``\emph{colorful}''$\in s_a(x)$,
and ``\emph{but}'', ``\emph{no}''$\in s_m(x)$.
Finally,
these extracted context words of each training instance will be included into $D$,
forming a final training corpus $D_s$ with
attention supervision information (\textbf{Lines 26-29}),
which will be used to conduct the last refined model training (\textbf{Line 30}).
The details will be provided in the next subsection.

\begin{table*}[t]			
\centering			
\small		
\resizebox{\textwidth}{18mm}{	
\begin{tabularx}{21cm}{|c|X|c|c|c|}			
\hline			
{\bf Iter} & \multicolumn{1}{|c|}{\bf Sentence} & {\bf Ans. / Pred.} & { \boldmath$E$ \unboldmath} & \boldmath {$x'_m$} \unboldmath \\			
\hline			
\hline			
{\bf 1} & \Appb{The}{0.1}\Appb{\bf [screen]}{0}\Appb{is}{0.2}\Appb{huge}{0.4}\Appb{and}{0.05}\Appb{colorful}{0.2}\Appb{but}{0.1}\Appb{no}{0.1}\Appb{led}{0.1}\Appb{back}{0.1}\Appb{\#\#light}{0.05}\Appb{\#\#ing}{0.1} & {\bf Pos / Pos} & {\bf 3.27} & {\bf \emph{huge}}\\			
\hline			
{\bf 2} & \Appb{The}{0.1}\Appb{\bf [screen]}{0}\Appb{is}{0.3}\Appb{$\langle mask\rangle$}{0}\Appb{and}{0.1}\Appb{colorful}{0.3}\Appb{but}{0.2}\Appb{no}{0.15}\Appb{led}{0.1}\Appb{back}{0.1}\Appb{\#\#light}{0.1}\Appb{\#\#ing}{0.1} & {\bf Pos / Pos} & {\bf 3.38} & {\bf \emph{is}}\\			
\hline			
{\bf 3} & \Appb{The}{0.15}\Appb{\bf [screen]}{0}\Appb{$\langle mask\rangle$}{0}\Appb{$\langle mask\rangle$}{0}\Appb{and}{0.1}\Appb{colorful}{0.4}\Appb{but}{0.2}\Appb{no}{0.2}\Appb{led}{0.15}\Appb{back}{0.1}\Appb{\#\#light}{0.1}\Appb{\#\#ing}{0.1} & {\bf Pos / Pos} & {\bf 3.31} & {\bf \emph{colorful}}\\		
\hline			
{\bf 4} & \Appb{The}{0.2}\Appb{\bf [screen]}{0}\Appb{$\langle mask\rangle$}{0}\Appb{$\langle mask\rangle$}{0}\Appb{and}{0.15}\Appb{$\langle mask\rangle$}{0}\Appb{but}{0.4}\Appb{no}{0.3}\Appb{led}{0.2}\Appb{back}{0.2}\Appb{\#\#light}{0.15}\Appb{\#\#ing}{0.2} & {\bf Pos / Neg} & {\bf 3.42} & {\bf \emph{but}}\\	
\hline			
{\bf 5} & \Appb{The}{0.15}\Appb{\bf [screen]}{0}\Appb{$\langle mask\rangle$}{0}\Appb{$\langle mask\rangle$}{0}\Appb{and}{0.2}\Appb{$\langle mask\rangle$}{0}\Appb{$\langle mask\rangle$}{0}\Appb{no}{0.4}\Appb{led}{0.25}\Appb{back}{0.25}\Appb{\#\#light}{0.2}\Appb{\#\#ing}{0.25} & {\bf Pos / Neg} & {\bf 3.24} & {\bf \emph{no}}\\	
\hline		
{\bf 6} & \Appb{The}{0.2}\Appb{\bf [screen]}{0}\Appb{$\langle mask\rangle$}{0}\Appb{$\langle mask\rangle$}{0}\Appb{and}{0.25}\Appb{$\langle mask\rangle$}{0}\Appb{$\langle mask\rangle$}{0}\Appb{$\langle mask\rangle$}{0}\Appb{led}{0.3}\Appb{back}{0.3}\Appb{\#\#light}{0.25}\Appb{\#\#ing}{0.3} & {\bf Pos / Neg} & {\bf 4.55} & {\bf \emph{---}}\\	
\hline		
\end{tabularx}}
\caption{\label{Table_Example2a}
The illustration of mining influential context words from the first training sentence in Table \ref{Table_Example1}.
$E(\alpha(x'))$ denotes the entropy of the \textbf{attention weight} distribution $\alpha(x')$,
$\epsilon_{\alpha}$ is entropy threshold set as $4.0$,
and $x'_m$ indicates the context word with the maximum attention weight.
Note that all extracted words are replaced with ``$\langle mask\rangle$'' and their background color is white.
}
\end{table*}

\begin{table*}[t]			
\centering			
\small			
\resizebox{\textwidth}{15mm}{
\begin{tabularx}{20.5cm}{|c|X|c|c|c|}			
\hline			
{\bf Iter} & \multicolumn{1}{|c|}{\bf Sentence} & {\bf Ans. / Pred.} & { \boldmath$E$ \unboldmath} & \boldmath {$x'_m$} \unboldmath \\			
\hline			
\hline			
{\bf 1} & \Appb{The}{0.05}\Appb{\bf [screen]}{0}\Appb{is}{0.2}\Appb{huge}{0.4}\Appb{and}{0.05}\Appb{colorful}{0.2}\Appb{but}{0.05}\Appb{no}{0.1}\Appb{led}{0.05}\Appb{back}{0.05}\Appb{\#\#light}{0.05}\Appb{\#\#ing}{0.1} & {\bf Pos / Pos} & {\bf 3.27} & {\bf \emph{huge}}\\			
\hline			
{\bf 2} & \Appb{The}{0.05}\Appb{\bf [screen]}{0}\Appb{is}{0.1}\Appb{$\langle mask\rangle$}{0}\Appb{and}{0.05}\Appb{colorful}{0.4}\Appb{but}{0.05}\Appb{no}{0.15}\Appb{led}{0.05}\Appb{back}{0.05}\Appb{\#\#light}{0.05}\Appb{\#\#ing}{0.1} & {\bf Pos / Pos} & {\bf 2.58} & {\bf \emph{colorful}}\\
\hline			
{\bf 3} & \Appb{The}{0.05}\Appb{\bf [screen]}{0}\Appb{is}{0.1}\Appb{$\langle mask\rangle$}{0}\Appb{and}{0.05}\Appb{$\langle mask\rangle$}{0}\Appb{but}{0.4}\Appb{no}{0.2}\Appb{led}{0.05}\Appb{back}{0.15}\Appb{\#\#light}{0.1}\Appb{\#\#ing}{0.15} & {\bf Pos / Pos} & {\bf 3.53} & {\bf \emph{but}}\\			
\hline			
{\bf 4} & \Appb{The}{0.1}\Appb{\bf [screen]}{0}\Appb{is}{0.15}\Appb{$\langle mask\rangle$}{0}\Appb{and}{0.15}\Appb{$\langle mask\rangle$}{0}\Appb{$\langle mask\rangle$}{0}\Appb{no}{0.4}\Appb{led}{0.2}\Appb{back}{0.15}\Appb{\#\#light}{0.1}\Appb{\#\#ing}{0.15} & {\bf Pos / Neg} & {\bf 3.55} & {\bf \emph{no}}\\		
\hline			
{\bf 5} & \Appb{The}{0.2}\Appb{\bf [screen]}{0}\Appb{is}{0.25}\Appb{$\langle mask\rangle$}{0}\Appb{and}{0.2}\Appb{$\langle mask\rangle$}{0}\Appb{$\langle mask\rangle$}{0}\Appb{$\langle mask\rangle$}{0}\Appb{led}{0.25}\Appb{back}{0.3}\Appb{\#\#light}{0.25}\Appb{\#\#ing}{0.3} & {\bf Pos / Neg} & {\bf 4.51} & {\bf \emph{---}}\\	
\hline
\end{tabularx}}
\caption{\label{Table_Example2b}
The illustration of mining influential context words from the first training sentence in Table \ref{Table_Example1}.
Here we define $E(\alpha(x'))$ as the entropy of the word saliency score distribution based on \textbf{partial gradients},
and
$\epsilon_{\alpha}$ is entropy threshold set as $4.0$,
}
\end{table*}

\subsection{Model Training with Attention Supervision Information} \label{SubSection_ModelTraining}

To exploit the extracted context words to refine the training of attention mechanisms for ABSA models,
we propose a soft attention regularizer
$\triangle(\alpha(s_a(x) \cup s_m(x)), \hat{\alpha}(s_a(x) \cup s_m(x)); \theta)$ to jointly optimize with the standard training objective.
Here $\alpha(*)$ and $\hat{\alpha}(*)$ respectively denote the model-induced and expected influence weight distributions of words in $s_a(x)\cup s_m(x)$,
and $\triangle(\alpha(*),\hat{\alpha}(*);\theta)$ is an \emph{Euclidean Distance} loss that penalizes the divergence between $\alpha(*)$ and $\hat{\alpha}(*)$.

As previously analyzed,
we expect to equally continue focusing on the context words of $s_a(x)$ during the final model training.
To this end,
we directly set their expected influence weights to the same value $\frac{1}{|s_a(x)|}$.
In this way,
the influence of words extracted first will be enhanced,
and that of words extracted later will be reduced,
avoiding the over-fitting of high-frequency context words with sentiment polarities and the under-fitting of low-frequency ones.
On the other hand,
for the words in $s_m(x)$ with misleading effects on the sentiment prediction of $x$,
we want to reduce their effects and thus directly set their expected weights as 0.
Back to the sentence shown in Table \ref{Table_Example2b},
both ``\emph{huge}'' and ``\emph{crowded}''$\in$$s_a(x)$ are assigned the same expected weight 0.5,
and both the expected weights of ``\emph{but}'' and ``\emph{no}''$\in$$s_m(x)$ are 0.

Finally,
our objective function on the training corpus $D_s$ with attention supervision information becomes
\begin{align}\label{Eqa_NewTrainingObjective}
&J_s(D_s;\theta) = -\sum_{(x,t,y)\in D_s} \{J(x,t,y;\theta)+ \\
&\gamma\triangle(\alpha(s_a(x)\cup s_m(x)),\hat{\alpha}(s_a(x)\cup s_m(x));\theta)\}, \notag
\end{align}where $J(x,t,y;\theta)$ is the conventional training objective defined in Equation \ref{Eqa_OldTrainingObjective},
and $\gamma$$>$$0$ is a hyper-parameter that balances the preference between the conventional loss function and the regularization term.
As a final note,
in addition to the utilization of attention supervision information,
our method is able to address the vanishing gradient problem
by adding such information into the intermediate layers of the entire network \citep{Szegedy:CVPR2015}. 
This is because the supervision of $\hat{\alpha}(*)$ is closer to $\alpha(*)$ than $y$.

\section{Experiments}\label{Section_Experiments}

To investigate the effectiveness of our approach,
we applied the proposed approach to improve MN, TNet-ATT and BERTABSA-ATT (See Section \ref{Section_Background})
and carried out a number of experiments on several benchmark datasets.

\subsection{Settings}
\textbf{Datasets}
The experimental datasets we used include
LAPTOP, REST \citep{Pontiki:SemEval2014} and TWITTER \citep{Dong:ACL2014},
where each sentence is paired with a given target aspect.
Table \ref{Table_Dataset} shows the statistics of these datasets.
Please note that we followed \citep{Chen:EMNLP2017} to remove a few instances with conflict sentiment labels.

\begin{table}[!t]
\small
\centering
\begin{tabular}{l|c|c|c|c}
\hline
\textbf{Domain} & \textbf{Dataset} & \textbf{\#Pos} & \textbf{\#Neg} & \textbf{\#Neu}\\
\hline
\hline
\multirow{2}*{LAPTOP}      & Train   & 980  & 858 & 454 \\
                           & Test    & 340  & 128 & 171 \\
\hline
\multirow{2}*{REST}        & Train   & 2159 & 800 & 632 \\
		                   & Test    & 730  & 195 & 196 \\
\hline
\multirow{2}*{TWITTER}     & Train   & 1567 & 1563& 3127\\
		                   & Test    & 174  & 174 & 346 \\
\hline
\end{tabular}
\caption{\label{Table_Dataset}
Datasets in our experiments.
\textbf{\#Pos}, \textbf{\#Neg} and \textbf{\#Neu} denote the numbers of instances with Positive, Negative and Neutral sentiment, respectively.
}
\end{table}

\textbf{Baselines}
We refer to our enhanced ABSA models that use attention weights to define $\alpha(x'_i)$ as MN(+AW-AS), TNet-ATT(+AW-AS) and BERTABSA-ATT(+A\ W-AS),
our enhanced ABSA models with $\alpha(x'_i)$ based on partial gradients as MN(+PG-AS), TNet-ATT(+PG-AS) and BERTABSA-ATT(+PG-AS), respectively.
Then,
we compared them with the following baselines:
\begin{itemize}
\item
MN, TNet, TNet-ATT, BERTABSA, and BERTABSA-ATT.
The standard contrast models described in Section \ref{Section_Background}.
Here, TNet-ATT and BERTABSA-ATT are the variants of TNet and BERTABSA, respectively,
equipped with attention mechanisms to generate aspect-related sentence representations.

\item
MN(+KT), \ \ \ \ TNet(+KT), \ \ \ \ TNet-ATT(+KT), \ \ \ \ BERTABSA(+KT), \ \ \ \ and BERTABSA-ATT(+KT).
The above-mentioned standard contrast models with additional $K$-iteration training,
which is also required in our enhanced models.

\item
MN(+Boosting), TNet-ATT(+Boosting) and BERTABSA-ATT(+Boosting).
To train such a contrast model,
we first individually train five baselines with different initializations,
which provides the final classification results via voting.

\item
MN(+Adaboost), TNet-ATT(+Adaboost) and BERTABSA-ATT(+Adaboo\ st).
Following \citep{Hastie:SIT2009},
we employ AdaBoost \citep{Freund:JCSS1997} to enhance our above-mentioned initial baselines.
Specifically,
starting with the unweighted training instances, we first build a classifier that produces class labels.
If a training instance is misclassified,
its weight in the training objective function is increased (boosted).
A second classifier is built using the new weights, which are no longer equal.
By continuously repeating the above training procedure for five iterations,
we can obtain five independent classifiers.
Finally,
we assign each classifier a score and define the final classifier as the linear combination of these five classifiers.
\end{itemize}

Besides,
to investigate the effects of different kinds of attention supervision information on our approach,
we report the performance of the following variants:
\begin{itemize}
\item
MN(+AW-AS$_a$), \ \ TNet-ATT(+AW-AS$_a$), \ \ BERTABSA-ATT(+AW-AS$_a$),
MN(+PG-AS$_a$), \ \ TNet-ATT(+PG-AS$_a$) and BERTABSA-ATT(+PG-AS$_a$).  
These variants only leverage the context words of $s_a(x)$ during the refined model training.

\item
MN(+AW-AS$_m$), \ TNet-ATT(+AW-AS$_m$), \ BERTABSA-ATT(+AW-AS$_m$),
MN(+PG-AS$_m$), TNet-ATT(+PG-AS$_m$) and BERTABSA-ATT(+PG-AS$_m$).
These variants only use the context words of $s_m(x)$ to train the refined models.
\end{itemize}

\begin{figure}[!t]
\centering
\subfigure[Effects of $\epsilon_{\alpha}$ on the validation sets with MN(+AW-AS)]{%
\label{Subfig_EAEffect_MNATT1}
\includegraphics[width=0.48\textwidth]{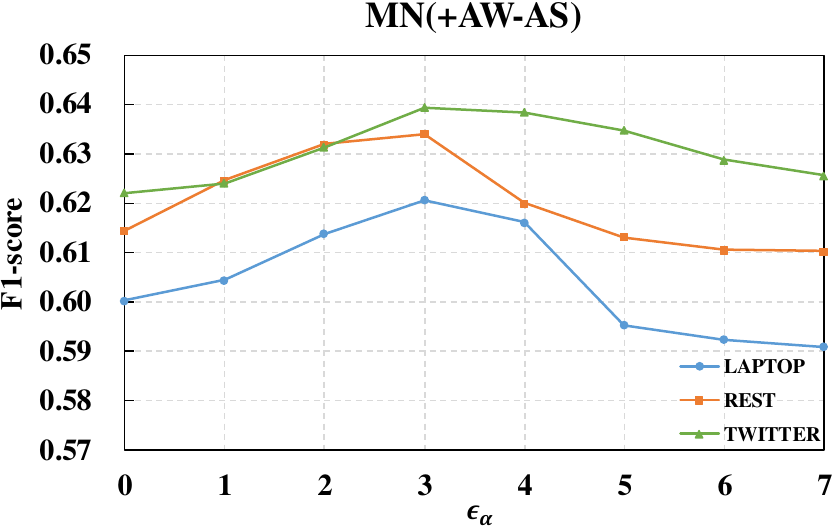}}
\subfigure[Effects of $\epsilon_{\alpha}$ on the validation sets with MN(+PG-AS)]{%
\label{Subfig_EAEffect_MNATT2}
\includegraphics[width=0.48\textwidth]{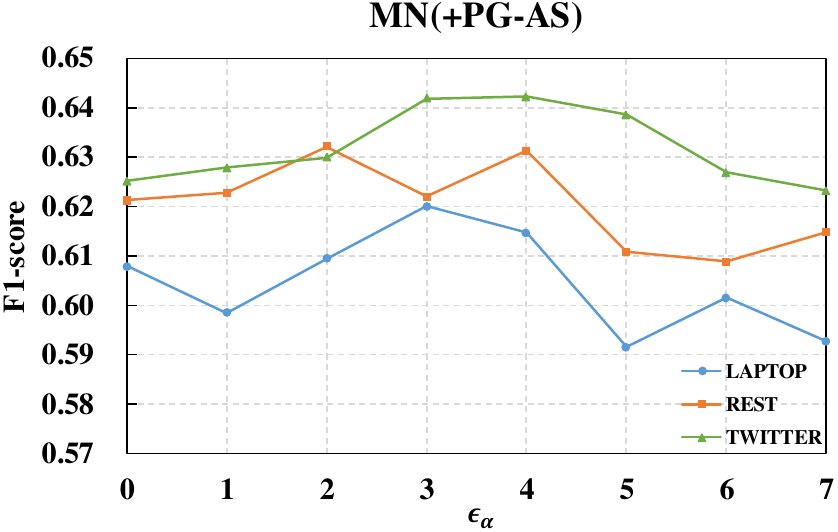}}
\subfigure[Effects of $\epsilon_{\alpha}$ on the validation sets with TNet-ATT(+AW-AS)]{%
\label{Subfig_EAEffect_TNetATT1}
\includegraphics[width=0.48\textwidth]{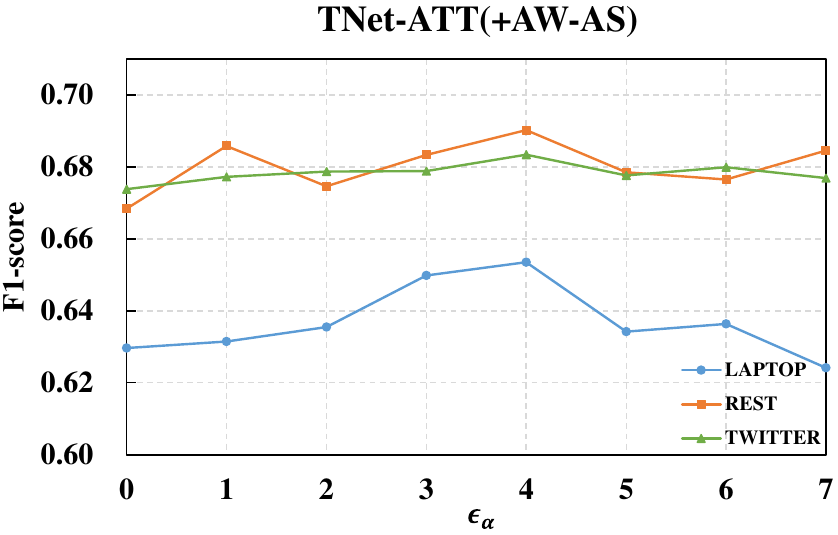}}
\subfigure[Effects of $\epsilon_{\alpha}$ on the validation sets with TNet-ATT(+PG-AS)]{%
\label{Subfig_EAEffect_TNetATT2}
\includegraphics[width=0.48\textwidth]{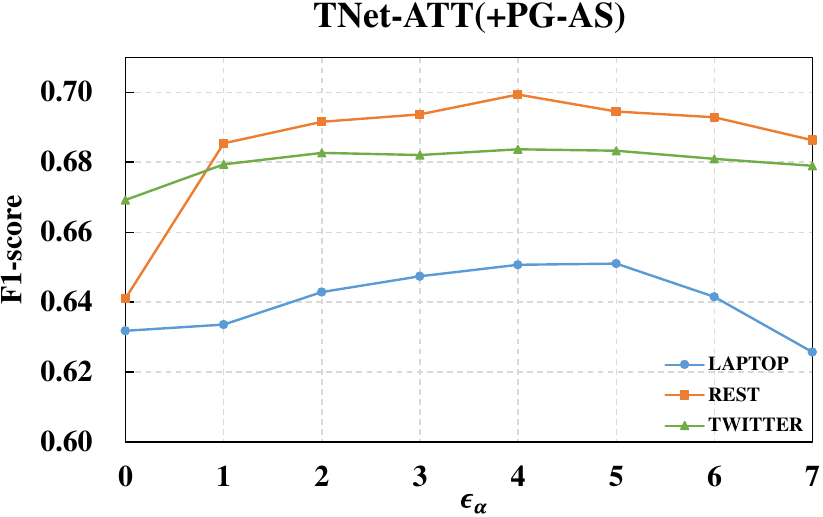}}
\subfigure[Effects of $\epsilon_{\alpha}$ on the validation sets with BERTABSA-ATT(+AW-AS)]{%
\label{Subfig_EAEffect_BERTABSAATT1}
\includegraphics[width=0.48\textwidth]{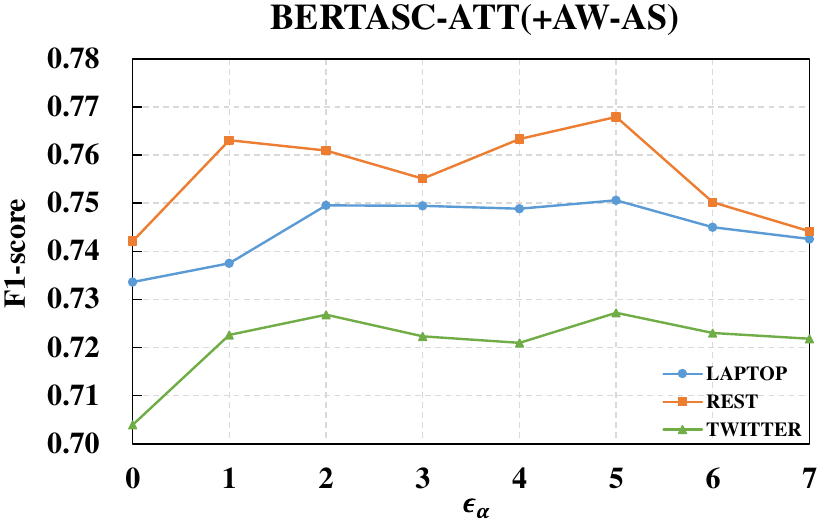}}
\subfigure[Effects of $\epsilon_{\alpha}$ on the validation sets with BERTABSA-ATT(+PG-AS)]{%
\label{Subfig_EAEffect_BERTABSAATT2}
\includegraphics[width=0.48\textwidth]{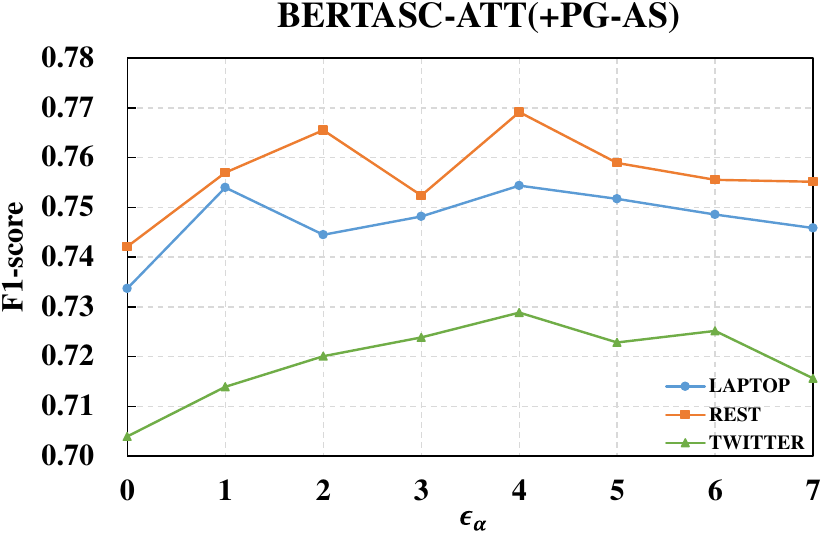}}
\caption{Experimental results for using attention weights to define $\alpha(x')$ (\ref{Subfig_EAEffect_MNATT1}, \ref{Subfig_EAEffect_TNetATT1}, \ref{Subfig_EAEffect_BERTABSAATT1}) and using the word saliency scores based on partial gradients to define $\alpha(x')$ (\ref{Subfig_EAEffect_MNATT2}, \ref{Subfig_EAEffect_TNetATT2}, \ref{Subfig_EAEffect_BERTABSAATT2}).}%
\label{Fig_DevExp}
\end{figure}

\textbf{Training Details}
For MN, TNet and their variants,
we used pre-trained \emph{GloVe} vectors \citep{Pennington:EMNLP2014} to initialize the word embeddings with the vector dimension 300.
In particular,
as implemented in \citep{Kim:EMNLP2014},
we randomly initialized the embeddings of out-of-vocabulary words according to the uniform distribution [-0.25,0.25].
Moreover,
we initialized other model parameters uniformly between [-0.01, 0.01].
During the model training,
we chose \emph{Adam} \citep{Kingma:ICLR2015} with the learning rate 0.001 as the optimizer.
To alleviate over-fitting,
we applied \emph{dropout} strategy \citep{Hinton:CS2012} with the rate 0.3 on the input word embeddings of the LSTM and the final aspect-related sentence representations.

For BERT and its variants, we used the uncased BERT-large pre-trained model to initialized the parameters.
During the model training, we chose \emph{Adam} with the learning rate 2e-5 as the optimizer.
The warm-up rate and dropout rate were all set to 0.1.

When implementing our approach,
we empirically set the maximum iteration number $K$ as 5,
$\gamma$ in Equation \ref{Eqa_NewTrainingObjective} as 0.1 on LAPTOP data set, 0.5 on REST data set and 0.1 on TWITTER data set, respectively.
All hyper-parameters were tuned on 20\% randomly extracted held-out training data.
Finally,
we used F1-Macro and accuracy as our evaluation metrics.

\subsection{Effects of $\epsilon_{\alpha}$}
From Line 14 of Algorithm \ref{Alg_OurApproach},
we observe that $\epsilon_{\alpha}$ is a crucial hyper-parameter that directly controls the amount of the extracted attention supervision information.
To investigate its effects on our approach,
we conducted experiments on the validation sets,
with $\epsilon_\alpha$ varying from 1.0 to 7.0.

\begin{table*}[!t]
\linespread{1.2}
\centering
\resizebox{\textwidth}{80mm}{
\begin{tabular}{l|c|c|c|c|c|c }
\hline
\multirow{2}*{\textbf{Model}} & \multicolumn{2}{c|}{\textbf{LAPTOP}} & \multicolumn{2}{c|}{\textbf{REST}} & \multicolumn{2}{c}{\textbf{TWITTER}} \\
\cline{2-7}
 & Macro-F1 & Accuracy & Macro-F1 & Accuracy & Macro-F1 & Accuracy \\
\hline
\hline
MN \citep{Wang:ACL2018}& 62.89 & 68.90 & 64.34 & 75.30 & --- & --- \\
MN                     & 63.28 & 68.97 & 65.88 & 77.32 & 66.17 & 67.71 \\
MN(+KT)                & 63.31 & 68.95 & 65.86 & 77.33 & 66.18 & 67.78 \\
MN(+Boosting)          & 64.17 & 69.28 & 66.23 & 77.66 & 67.12 & 68.14 \\
MN(+Adaboost)          & 60.52 & 67.88 & 62.29 & 76.77 & 65.09 & 66.96 \\
\hline
MN(+AW-AS$_m$)         & 64.37 & 69.69 & 68.40 & 78.13 & 67.20 & 68.90 \\
MN(+AW-AS$_a$)         & 64.61 & 69.95 & 68.59 & 78.23 & 67.47 & 69.17 \\
MN(+AW-AS)             & 65.24 & 70.53 & 69.15 & 78.75 & \textbf{67.88}$^{**}$ & 69.64 \\
MN(+PG-AS$_m$)         & 64.66 & 70.06 & 68.43 & 78.19 & 66.87 & 68.99\\
MN(+PG-AS$_a$)         & 64.93 & 70.22 & 68.89 & 78.27 & 67.13 & 69.60 \\
MN(+PG-AS)             & \textbf{65.58}$^{**}$ & \textbf{70.84}$^{**}$ & \textbf{69.42}$^{**}$ & \textbf{78.98}$^{**}$ & 67.80 & \textbf{69.78}$^{**}$ \\
\hline
\hline
TNet                   & 67.75 & 72.67 & 69.53 & 78.98 & 70.97 & 72.74 \\
TNet(+KT)              & 67.87 & 72.46 & 68.42 & 78.71 & 71.54 & 73.12 \\
TNet-ATT               & 67.13 & 72.56 & 67.71 & 79.31 & 70.23 & 71.67 \\
TNet-ATT(+KT)          & 66.79 & 72.46 & 68.33 & 79.52 & 70.03 & 72.25 \\
TNet-ATT(+Boosting)    & 68.16 & 73.40 & 69.28 & \textbf{80.14}$^{**}$ & 71.33 & 72.98 \\
TNet-ATT(+Adaboost)    & 63.73 & 69.33 & 62.95 & 75.85 & 66.26 & 68.35 \\
\hline
TNet-ATT(+AW-AS$_m$)   & 68.03 & 73.01 & 69.85 & 78.99 & 71.00 & 72.79 \\
TNet-ATT(+AW-AS$_a$)   & 69.08 & 73.70 & 70.51 & 80.10 & 71.19 & 73.06 \\
TNet-ATT(+AW-AS)       & 69.28 & 74.18 & 70.79 & 80.11 & 71.44 & 73.12 \\
TNet-ATT(+PG-AS$_m$)   & 68.35 & 73.36 & 71.16 & 79.25 & 71.24 & 72.86 \\
TNet-ATT(+PG-AS$_a$)   & 69.41 & 74.37 & 71.22 & 79.88 & 71.64 & 73.05 \\
TNet-ATT(+PG-AS)       & \textbf{69.67}$^{**}$ & \textbf{74.49}$^{**}$ & \textbf{71.72}$^{**}$ & \textbf{80.14}$^{**}$ & \textbf{71.99}$^{**}$ & \textbf{73.41}$^{**}$ \\
\hline
\hline
BERT-based ABSA \citep{Hu:ACL2019}      & ---   & 81.39 & ---   & ---   & --- & --- \\
BERT-based ABSA \citep{Song:Arxiv2019}  & 76.31 & 79.39 & 73.76 & 83.12 & 73.13 & 74.71 \\
BERT-based ABSA \citep{Ke:Arxiv2019}    & 76.47 & 80.72 & 79.20 & 86.16 & --- & --- \\
BERTABSA                 & 77.88 & 81.38 & 80.49 & 86.61 & 75.67 & 76.59 \\
BERTABSA(+KT)            & 77.98 & 81.38 & 80.43 & 86.78 & 75.78 & 76.80 \\
BERTABSA-ATT             & 78.02 & 81.53 & 80.71 & 86.79 & 74.98 & 76.25 \\
BERTABSA-ATT(+KT)        & 78.05 & 81.69 & 80.59 & 86.85 & 75.05 & 76.34 \\
BERTABSA-ATT(+Boosting)          & 78.15 & 81.60 & 81.67 & 87.59 & 76.05 & 76.99 \\
BERTABSA-ATT(+Adaboost)          & 77.08 & 80.85 & 78.52 & 85.89 & 74.06 & 75.19 \\
\hline
BERTABSA-ATT(+AW-AS$_m$) & 78.21 & 81.82 & 80.98 & 87.02 & 75.23 & 76.52 \\
BERTABSA-ATT(+AW-AS$_a$) & 78.97 & 82.10 & 81.47 & 87.39 & 75.92 & 77.06 \\
BERTABSA-ATT(+AW-AS)     & 79.06 & 82.21 & 81.65 & 87.45 & 76.19 & 77.21 \\
BERTABSA-ATT(+PG-AS$_m$) & 78.19 & 81.84 & 81.06 & 87.09 & 75.20 & 76.41 \\
BERTABSA-ATT(+PG-AS$_a$) & 79.16 & 82.47 & 82.21 & 87.64 & 76.33 & 77.48 \\
BERTABSA-ATT(+PG-AS)     & \textbf{79.30}$^{**}$ & \textbf{82.64}$^{**}$ & \textbf{82.34}$^{**}$ & \textbf{87.86}$^{**}$ & \textbf{76.45}$^{**}$ & \textbf{77.60}$^{**}$ \\
\hline
\end{tabular}}
\caption{\label{Table_OverallResults}
Experimental results on various datasets.
We directly report the best experimental results of MN and TNet in \citep{Wang:ACL2018,Li:ACL2018}.
$\ast\ast$ and $\ast$ denote statistical significance at $p<$0.01 and $p<$0.05 over the baselines (MN, TNet, BERTABSA) on each test set, respectively.
Here we conducted 1,000 bootstrap tests \citep{Koehn:EMNLP2004} to measure the significance in metric score differences.
}
\end{table*}

Experimental results of different enhanced models on the validation sets are provided in Figure \ref{Fig_DevExp}.
When using attention weights to define $\alpha(x')$,
MN(+AS), ATT(+AS) and BERTABSA(+AS) achieve the best performance with $\epsilon_{\alpha}$ being set to 3.0, 4.0 and 5.0, respectively.
Meanwhile,
if we define $\alpha(x')$ based on partial gradients,
the optimal $\epsilon_{\alpha}$s for MN(+AS), ATT(+AS) and BERTABSA(+AS) are 3.0, 4.0 and 4.0, respectively.
Therefore,
we used the above fixed optimal $\epsilon_{\alpha}$s for different ABSA models in the following experiments.

\subsection{Overall Results}
Table \ref{Table_OverallResults} provides the overall experimental results.
For a comprehensive comparison, we also display the previously reported scores of MN \citep{Wang:ACL2018} and TNet \citep{Li:ACL2018} on the same data set.
According to the experimental results,
we can reach the following conclusions:

\textbf{First},
both of our reimplemented MN and TNet are comparable to their original models reported in \citep{Wang:ACL2018,Li:ACL2018}.
Besides,
our BERT based ABSA model also exhibits slightly better performance than \citep{Hu:ACL2019,Song:Arxiv2019,Ke:Arxiv2019} that are also based on BERT.
These results indicate that our reimplemented baselines are competitive.
When we replace the CNN layer of TNet with an attention mechanism,
TNet-ATT is slightly inferior to TNet.
Even we perform additional $K$-iterations of training on these models,
their performance has not changed significantly.
All these results suggest that simply increasing training time is unable to enhance the performance of neural ABSA models.
\footnote{The performance of TNet reported in our paper is lower than the original paper \citep{Li:ACL2018} because the authors use an incorrect data pre-processing step in their released code: https://github.com/lixin4ever/TNet.
Specifically, they use \emph{t.strip()} to strip sentiment tags from target words which may cause an error. For example, the output of \emph{`nicki/n'.strip(`/n')} is ``\emph{icki}'' rather than ``\emph{nicki}''.
However, even using the incorrectly pre-processed datasets, our models can still outperform the baselines.
The related results are provided in Appendix.}

\textbf{Second},
when we apply the proposed approach to MN, TNet-ATT and BERT-ATT,
the performance of all these models is significantly enhanced.
Particularly,
on the whole,
the context words in $s_a(x)$ are more effective than those in $s_m(x)$.
This is because the proportion of correctly predicted training instances is larger than that of incorrectly predicted instances.
Besides,
the performance gaps between MN(+AW-AS$_a$) and MN(+AW-AS$_m$), MN(+PG-AS$_a$) and MN(+PG-AS$_m$) are smaller than those between two variants of TNet-ATT, two variants of BERT-ATT.
One underlying reason is that the performance of both TNet-ATT and BERT-ATT is better than MN,
which enables TNet-ATT, BERT-ATT to produce more correctly predicted training instances.
This in turn brings more attention supervision to TNet-ATT, BERT-ATT than MN.

\textbf{Third},
when we use both kinds of attention supervision information ($s_a(x)$ and $s_m(x)$),
no matter for which metric,
both MN(+AW-AS) and MN(+PG-AS) remarkably outperform MN on all test sets.
Although our TNet-ATT is slightly inferior to TNet,
both TNet-ATT(+AW-AS) and TNet-ATT(+PG-AS) still significantly surpasses both TNet and TNet-ATT.
Similarly,
the utilization of the attention mechanism lead to the slight performance decline of BERT-ATT,
however, both BERT-ATT(+AW-AS) and BERT-ATT(+PG-AS) still have better performance than BERT-ATT.
Even compared with the enhanced models using boosting or Adaboost, 
our enhanced models based on partial gradients always exhibit better performance.
These results strongly demonstrate the effectiveness and generality of our approach.

\textbf{Finally},
as for two definitions of $\alpha(x_i)$,
the definition based on partial gradients is a better choice to identify useful attention supervision information, leading to more significant improvements for all models.
These results echo with the experimental results in other tasks \citep{Jain:NAACL2019, Serrano:ACL2019, Ding:WMT2019}.
Therefore,
all the following experiments only consider defining $\alpha(x_i)$ based on partial gradients.

\subsection{Our Approach vs. Randomly Masking}
Using our approach,
at each iteration,
we mask the most influential context word to shield its effect during the subsequent process of extracting attention supervision information.
To demonstrate the validity of such operation,
we also reported the performance of enhanced models,
where the context words are randomly masked during the process of extracting attention supervision information.

Table \ref{Table_ESIResults} displays the experimental results.
We can observe that
randomly masking significantly degrades the performance of all enhanced models.
These results strongly show that our such an operation is indeed beneficial to extract useful attention supervision information.

Notably, it can be seen that the randomly masked model can still slightly outperform the baseline.
It might be because randomly masking can be regarded as a special dropout or noise, which prevents the model from overfitting.

\begin{table*}[t]
\renewcommand
\baselinestretch{1.1}
\centering
\resizebox{\textwidth}{20mm}{
\begin{tabular}{l|c|c|c|c|c|c }
\hline
\multirow{2}*{\textbf{Model}} & \multicolumn{2}{c|}{\textbf{LAPTOP}} & \multicolumn{2}{c|}{\textbf{REST}} & \multicolumn{2}{c}{\textbf{TWITTER}} \\
\cline{2-7}
 & Macro-F1 & Accuracy & Macro-F1 & Accuracy & Macro-F1 & Accuracy \\
\hline
\hline
BERTABSA                & 77.88 & 81.38 & 80.49 & 86.61 & 75.67 & 76.59 \\
BERTABSA-ATT            & 78.02 & 81.53 & 80.71 & 86.79 & 74.98 & 76.25 \\
BERTABSA-ATT(+Boosting) & 78.96 & 82.16 & 81.67 & 87.59 & 76.05 & 76.99 \\
BERTABSA-ATT(+Adaboost) & 78.04 & 81.48 & 80.24 & 86.76 & 74.57 & 76.00\\
\hline
BERTABSA-ATT(+RM-AS)    & 78.29 & 81.80 & 80.95 & 87.06 & 75.71 & 76.89 \\
BERTABSA-ATT(+AW-AS)    & 79.06 & 82.21 & 81.65 & 87.45 & 76.19 & 77.21 \\
BERTABSA-ATT(+PG-AS)    & \textbf{79.30} & \textbf{82.64} & \textbf{82.34} & \textbf{87.86} & \textbf{76.45} & \textbf{77.60} \\
\hline
\end{tabular}}
\caption{
\label{Table_ESIResults}
Experimental results of using different approaches to extract supervision information.
BERTABSA-ATT(+RM-AS) is an ABSA model that employs randomly masking operation to extract supervision information.
}
\end{table*}

\begin{table*}[t]
\renewcommand
\baselinestretch{1.1}
\centering
\resizebox{\textwidth}{48mm}{
\begin{tabular}{c|c|c|c|c|c|c|c }
\hline
\textbf{Training} & \multirow{2}*{\textbf{Model}} & \multicolumn{2}{c|}{\textbf{LAPTOP}} & \multicolumn{2}{c|}{\textbf{REST}} & \multicolumn{2}{c}{\textbf{TWITTER}} \\
\cline{3-8}
\textbf{Corpus Size} & & Macro-F1 & Accuracy & Macro-F1 & Accuracy & Macro-F1 & Accuracy \\
\hline
\hline
\multirow{6}*{25\%} & BERTABSA                & 77.15 & 80.49 & 75.29 & 84.20 & 69.72 & 71.56 \\
                    & BERTABSA-ATT            & 75.79 & 79.59 & 76.11 & 84.08 & 70.77 & 72.54 \\
                    & BERTABSA-ATT(+Boosting) & 76.52 & 79.71 & 77.49 & 84.90 & 70.96 & 72.50 \\
                    & BERTABSA-ATT(+Adaboost) & 76.16 & 79.51 & 77.04 & 84.61 & 69.81 & 71.76 \\
                    & BERTABSA-ATT(+AW-AS)    & 77.04 & 79.89 & 75.99 & 84.08 & 71.12 & 72.40 \\
                    & BERTABSA-ATT(+PG-AS)    & \textbf{77.35} & \textbf{80.99} & \textbf{77.52} & \textbf{85.15} & \textbf{72.15} & \textbf{73.27} \\
\hline
\hline
\multirow{6}*{50\%} & BERTABSA                & 77.25 & 80.59 & 77.65 & 84.91 & 73.23 & 74.29 \\
                    & BERTABSA-ATT            & 76.43 & 80.05 & 76.94 & 84.80 & 72.58 & 74.28 \\
                    & BERTABSA-ATT(+Boosting) & 76.97 & 80.38 & 77.77 & 85.12 & 72.83 & 74.32 \\
                    & BERTABSA-ATT(+Adaboost) & 76.34 & 79.95 & 77.90 & 85.09 & 72.42 & 73.79 \\
                    & BERTABSA-ATT(+AW-AS)    & 77.24 & 79.89 & 77.56 & 85.74 & 73.20 & 74.42 \\
                    & BERTABSA-ATT(+PG-AS)    & \textbf{78.80} & \textbf{81.46} & \textbf{78.98} & \textbf{85.96} & \textbf{73.30} & \textbf{74.57} \\
\hline
\hline
\multirow{6}*{75\%} & BERTABSA                & 77.45 & 80.91 & 79.05 & 86.17 & 73.92 & 75.48 \\
                    & BERTABSA-ATT            & 77.90 & 81.38 & 79.14 & 85.78 & 74.32 & 75.58 \\
                    & BERTABSA-ATT(+Boosting) & 77.93 & 82.00 & 79.28 & 86.11 & 75.08 & 76.16 \\
                    & BERTABSA-ATT(+Adaboost) & 77.71 & 81.15 & 79.17 & 85.82 & 74.10 & 75.43 \\
                    & BERTABSA-ATT(+AW-AS)    & \textbf{78.91} & \textbf{82.09} & 80.03 & 86.23 & 74.21 & 75.72 \\
                    & BERTABSA-ATT(+PG-AS)    & 78.66 & 82.04 & \textbf{80.53} & \textbf{86.63} & \textbf{75.47} & \textbf{76.52} \\
\hline
\hline
\multirow{6}*{100\%} & BERTABSA                & 77.88 & 81.38 & 80.49 & 86.61 & 75.67 & 76.59 \\
                     & BERTABSA-ATT            & 78.02 & 81.53 & 80.71 & 86.79 & 74.98 & 76.25 \\
                     & BERTABSA-ATT(+Boosting) & 78.96 & 82.16 & 81.67 & 87.59 & 76.05 & 76.99 \\
                     & BERTABSA-ATT(+Adaboost) & 78.04 & 81.48 & 80.24 & 86.76 & 74.57 & 76.00\\
                     & BERTABSA-ATT(+AW-AS)    & 79.06 & 82.21 & 81.65 & 87.45 & 76.19 & 77.21 \\
                     & BERTABSA-ATT(+PG-AS)    & \textbf{79.30} & \textbf{82.64} & \textbf{82.34} & \textbf{87.86} & \textbf{76.45} & \textbf{77.60} \\
\hline
\end{tabular}}
\caption{
\label{Table_ETCSResults}
Experimental results of using different sizes of training corpora.
}
\end{table*}

\subsection{Effect of Training Corpus Size}
To investigate the generality of our approach,
we conducted experiments with different sizes of training corpora.

From the experimental results reported in Table \ref{Table_ETCSResults},
we observe that no matter how large training data we used,
our enhanced models consistently outperform their corresponding baseline models.
Therefore,
we confirm that our approach is general to these neural ABSA models with different sizes of training corpora.

\begin{table*}[t]
\renewcommand
\baselinestretch{1.1}
\centering
\resizebox{0.9\textwidth}{30mm}{
\begin{tabular}{l|p{18mm}<{\centering}|p{18mm}<{\centering}|p{18mm}<{\centering}|p{18mm}<{\centering}|p{18mm}<{\centering}|p{18mm}<{\centering}}
\hline
\multirow{2}*{\textbf{Model}} & \multicolumn{2}{c|}{\textbf{LAPTOP}} & \multicolumn{2}{c|}{\textbf{REST}} & \multicolumn{2}{c}{\textbf{TWITTER}} \\
\cline{2-7}
 & \multicolumn{1}{c|}{Train} & Test & \multicolumn{1}{c|}{Train} & Test & \multicolumn{1}{c|}{Train} & Test \\
\hline
\hline
BERTABSA                & 627.09 & 6.56 & 1290.53 & 11.30 & 1666.43 & 7.26 \\
BERTABSA (+KT)          & 3135.45 & 6.53 & 6452.65 & 11.51 & 8332.15 & 7.34 \\
BERTABSA-ATT            & 718.32 & 6.36 & 1357.65 & 11.72 & 1769.22 & 7.33 \\
BERTABSA-ATT (+KT)      & 3609.40 & 6.49 & 6797.75 & 11.68 & 8853.98 & 7.35 \\
\hline
BERTABSA-ATT(+Boosting) & 3598.12 & 31.94 & 6788.53 & 60.28 & 8832.12 & 36.71 \\
\qquad each baseline    & 719.76 & --- & 1358.77 & --- & 1780.08 & --- \\
BERTABSA-ATT(+Adaboost) & 3614.84 & 30.87 & 6810.22 & 61.32 & 8928.83 & 37.82 \\
\qquad each iteration   & 723.46 & --- & 1363.78 & --- & 1777.36 & --- \\
\hline
BERTABSA-ATT(+AW-AS)    & 4,332.37 & 6.43 & 8,159.79 & 11.51 & 10,639.75 & 7.45 \\
\qquad each mining iteration   & 721.88 & --- & 1,359.55 & --- & 1,770.80 & --- \\
\qquad the final model training      & 722.97 & --- & 1,362.04 & --- & 1,785.77 & --- \\
BERTABSA-ATT(+PG-AS)    & 11,667.97 & 6.49 & 19,822.11 & 11.93 & 31,283.85 & 7.22 \\
\qquad each mining iteration   & 2,190.09 & --- & 3,691.99 & --- & 5,905.72 & --- \\
\qquad the final model training      & 717.53 & --- & 1,362.04 & --- & 1,755.26 & --- \\
\hline
\end{tabular}}
\caption{
\label{Table_Speed}
The time (seconds) required for training and testing.
}
\end{table*}

\subsection{Model Speed}
To investigate the training and testing efficiency of various ABSA models, we show their training and testing speeds on a NVIDIA GeForce GTX TITAN X GPU in Table \ref{Table_Speed}.
Because our approach requires additional 5 iterations to automatically mine supervision information for attention mechanisms, BERTABSA-ATT(+AW-AS) approximately takes 6 times as long as the baseline model, which is comparable with BERTABSA (+KT), BERTABSA-ATT (+KT), BERTABSA-ATT(+Boosting) and BERTABSA-ATT(+Adaboost).
By comparison, BERTABSA-ATT(+PG-AS) consumes more time to extract gradient-based supervision information due to the random sampling disturbance (as described in Definition 2 of Subsection \ref{SubSection_ExtractAttentionSupervision}).

Please note that our approach only affects the training optimization of neural ABSA models, 
without any impact on the model testing.
Since the training can be done offline, we believe that the training time is not critical to real-world application as the online classifications on instances is very fast.

\subsection{Progressive Self-supervision for Self-attention within the BERT}
Intuitively, we can directly apply progressive self-supervised attention learning to enhance the learning of self-attentions within the BERT model itself, so as to avoid introducing the additional attention mechanism on top of BERT. 
However, our method is not directly suitable for the learning of the BERT self-attention mechanism, due to the following two reasons: 
(1) The BERT self-attention mechanism uses multiple heads
\footnote{Besides, the BERT self-attention mechanism is with multiple layers. 
Thus, choosing a suitable layer is also difficult. 
Considering the computational cost, we only impose our approach on the top layer of the BERT self-attention mechanism.}, 
and thus it is difficult to decide which attention head is to be imposed on by our approach; 
(2) The outputs of the self-attention mechanism are mainly word-level representations, which cannot be directly used to conduct the final classification. 

To deal with the above-mentioned issues,
we explore three approaches to directly refining the self-attention mechanism in BERTABSA (See Section \ref{SubSection_BERTbasedABSAModel}):
(1) We randomly select one attention head, on which we impose our proposed approach;
(2) For each attention head, we first calculate the gradient-based saliency scores of words, 
\footnote{Previous experiments in this paper show that the gradient-based saliency score performs better than the attention weight-based saliency score. 
Therefore, we mainly focus on the approaches involving gradient-based saliency scores in this group of experiments.}
which are based on the partial gradients of the word representations with respect to the representations of aspect words. (See Section \ref{SubSection_ExtractAttentionSupervision}). 
Then, we calculate the attention head-specific entropy of the normalized saliency scores.
Finally, we employ our method to only refine the attention head with the smallest entropy;
(3) As implemented in (2), for each word, we calculate its saliency score specific to the attention head, and then define its final saliency score as the average score of all heads. 
Lastly, via a soft attention regularizer, we use the obtained attention supervised information to constrain the average attention weight during the model training.

Besides,
we explore another kind of BERT-based ABSA model: BERTABSA-CLS.
Unlike BERTABSA, 
this model is fed with the input: ``[CLS] + Sentence + [SEP] + Aspect'', where the representation of [CLS] is used to conduct the final classification. 
Likewise,
we employ the above-mentioned three approaches to refine the self-attention mechanism of BERTABSA-CLS,
with the only difference that the word-level saliency scores are calculated according to the average partial gradients of the word representations with respect to the CLS representation.

\begin{table*}[t]
\renewcommand
\baselinestretch{1.1}
\centering
\resizebox{\textwidth}{24mm}{
\begin{tabular}{l|c|c|c|c|c|c|c }
\hline
\multirow{2}*{\textbf{Model}} & \multirow{2}*{\textbf{Attention Head}} & \multicolumn{2}{c|}{\textbf{LAPTOP}} & \multicolumn{2}{c|}{\textbf{REST}} & \multicolumn{2}{c}{\textbf{TWITTER}} \\
\cline{3-8}
 & & Macro-F1 & Accuracy & Macro-F1 & Accuracy & Macro-F1 & Accuracy \\
\hline
\hline
BERTABSA-ATT            & --- & 78.02 & 81.53 & 80.71 & 86.79 & 74.98 & 76.25 \\
\hline
BERTABSA-ATT(+PG-AS)    & --- & \textbf{79.30} & \textbf{82.64} & \textbf{82.34} & \textbf{87.86} & \textbf{76.45} & \textbf{77.60} \\
\hline
\hline
BERTABSA                & --- & 77.88 & 81.38 & 80.49 & 86.61 & 75.67 & 76.59 \\
\hline
\multirow{3}*{BERTABSA(PG-AS)} & Randomly Select & 78.56 & 81.69 & 80.69 & 86.85 & 75.70 & 76.95 \\
                               & Smallest Entropy & 78.45 & 81.85 & 81.05 & 86.94 & 76.25 & 77.24 \\
                               & Average & 78.90 & 82.26 & 81.60 & 87.39 & 76.01 & 77.12 \\
\hline
\hline
BERTABSA-CLS            & --- & 76.98 & 80.56 & 80.21 & 86.72 & 75.10 & 76.13 \\
\hline
\multirow{3}*{BERTABSA-CLS(PG-AS)}    & Randomly Select & 77.18 & 80.54 & 80.50 & 87.03 & 75.79 & 76.73 \\
                                      & Smallest Entropy & 77.30 & 80.66 & 80.52 & 86.85 & 75.43 & 76.62 \\
                                      & Average & 78.01 & 81.01 & 80.25 & 87.03 & 74.98 & 76.35 \\
\hline
\end{tabular}}
\caption{
\label{Table_SelfAttention}
Experimental results of using our method to enhance the learning of self-attentions within the BERT model itself.
}
\end{table*}

Table \ref{Table_SelfAttention} reports the final results. 
We can observe that using the above-mentioned approaches, the performance of both BERTABSA and BERTABSA-CLS can be slightly improved, which, however, is still obviously inferior to BERTABSA-ATT(+PG-AS). 
For these results, we speculate that BERT is mainly used to learn contextualized representations independent from specific tasks. 
Stacking an attention mechanism on top of BERT, which has no effect on the integrity of BERT, is a better choice than using classification task-specific signals to guide the self-attention learning within BERT.

\begin{table*}[t]
\centering
\resizebox{\textwidth}{16mm}{
\begin{tabularx}{24.5cm}{|l|X|c|c|c|}
\hline
{\bf Model} & \multicolumn{1}{|c|}{\bf Sentence} & {\bf Ans./Pred.}
\unboldmath\\
\hline
\hline
{\bf BERTABSA} & \Appb{\bf [Price]}{0}\Appb{was}{0.2}\Appb{higher}{0.25}\Appb{when}{0.2}\Appb{purchased}{0.1}\Appb{on}{0.05}\Appb{mac}{0.05}\Appb{when}{0.05}\Appb{compared}{0.05}\Appb{to}{0.05}\Appb{price}{0.1}\Appb{showing}{0.1}\Appb{on}{0.05}\Appb{PC}{0.05}\Appb{...}{0.04}& {\bf Neg / Pos}\\
\hline
{\bf BERTABSA-ATT} & \Appb{\bf [Price]}{0}\Appb{was}{0.15}\Appb{higher}{0.4}\Appb{when}{0.05}\Appb{purchased}{0.1}\Appb{on}{0.05}\Appb{mac}{0.05}\Appb{when}{0.1}\Appb{compared}{0.1}\Appb{to}{0.05}\Appb{price}{0.1}\Appb{showing}{0.05}\Appb{on}{0.05}\Appb{PC}{0.05}\Appb{...}{0.04}& {\bf Neg / Neg}\\
\hline
{\bf BERTABSA-ATT(+PG-AS)} & \Appb{\bf [Price]}{0}\Appb{was}{0.25}\Appb{higher}{0.4}\Appb{when}{0.1}\Appb{purchased}{0.05}\Appb{on}{0.05}\Appb{mac}{0.05}\Appb{when}{0.1}\Appb{compared}{0.1}\Appb{to}{0.05}\Appb{price}{0.1}\Appb{showing}{0.05}\Appb{on}{0.05}\Appb{PC}{0.05}\Appb{...}{0.04}& {\bf Neg / Neg}\\
\hline
\hline
{\bf BERTABSA} & \Appb{Not}{0.1}\Appb{as}{0.1}\Appb{fast}{0.1}\Appb{as}{0.1}\Appb{I}{0.1}\Appb{would}{0.1}\Appb{have}{0.1}\Appb{expect}{0.1}\Appb{for}{0.1}\Appb{an}{0.25}\Appb{\bf [I}{0}\Appb{\bf 5]}{0}\Appb{.}{0.1}& {\bf Neg / Pos}\\
\hline
{\bf BERTABSA-ATT} & \Appb{Not}{0.05}\Appb{as}{0.1}\Appb{fast}{0.05}\Appb{as}{0.05}\Appb{I}{0.05}\Appb{would}{0.05}\Appb{have}{0.05}\Appb{expect}{0.05}\Appb{for}{0.05}\Appb{an}{0.45}\Appb{\bf [I}{0}\Appb{\bf 5]}{0}\Appb{.}{0.05}& {\bf Neg / Pos}\\
\hline
{\bf BERTABSA-ATT(+PG-AS)} & \Appb{Not}{0.25}\Appb{as}{0.1}\Appb{fast}{0.35}\Appb{as}{0.1}\Appb{I}{0.1}\Appb{would}{0.1}\Appb{have}{0.1}\Appb{expect}{0.1}\Appb{for}{0.15}\Appb{an}{0.1}\Appb{\bf [I}{0}\Appb{\bf 5]}{0}\Appb{.}{0.1}& {\bf Neg / Neg}\\
\hline
\end{tabularx}}
\caption{\label{Table_Example3}
Two instances predicted by the three BERT-based ABSA models.
}
\end{table*}

\subsection{Case Study}
To know how our method improves neural ABSA models,
we deeply analyzed the attention results of BERTABSA-ATT and BERTABSA-ATT(+PG-AS).
It has been found that our proposed approach can solve the above-mentioned two issues well.

Table \ref{Table_Example3} provides two examples.
BERTABSA incorrectly predicts the sentiment of the first test sentence as positive.
This is because the context word ``\emph{higher}'' only appears in few training instances with negative polarities.
Therefore,
BERTABSA does not fully exploit the sentiment information encoded by ``\emph{higher}''.
When using BERTABSA-ATT and BERTABSA-ATT(+PG-AS),
the attention weight of ``\emph{higher}'' is increased than that of baseline in training instances.
Thus,
both BERTABSA-ATT and BERTABSA-ATT(+PG-AS) are capable of assigning a greater attention weight
(0.263$\rightarrow$0.636, 0.391) to this context word,
leading to the correct predictions of the first test sentence.
For the second test sentence,
both BERTABSA and BERTABSA-ATT incorrectly pay much attention to the context word ``\emph{an}''
while ignoring the other two context words ``\emph{not}'' and ``\emph{fast}''.
By contrast,
only BERTABSA-ATT(+PG-AS) can assign greater attention weights to ``\emph{not}'' (0.072, 0.005 $\rightarrow$ 0.113) and ``\emph{fast}'' (0.087, 0.005 $\rightarrow$ 0.376).
Thus,
it can avoid the noisy effect of ``\emph{an}'' and make a correct prediction.

\section{Conclusion and Future Work}
In this paper,
we have explored automatically mining supervision information for attention mechanisms of neural ABSA models.
Through in-depth analyses,
we first reveal the defect of the attention mechanism for ABSA:
a few frequent words with sentiment polarities tend to be over-learned,
while those with low frequencies often lack sufficient learning.
Then,
we propose a novel approach to automatically and incrementally mine attention supervision information for neural ABSA models.
The mined information can be further used to refine the model training via a regularization term.
To verify the effectiveness of our approach,
we apply our approach to three dominant neural ABSA models.
Experimental results demonstrate our method significantly improves
the performance of these models.

In the future,
we plan to extend our approach to other neural NLP tasks with attention mechanisms,
such as neural document classification \citep{Yang:NAACL2016} and neural machine translation \citep{Zhang:TPAMI2018}.

\section*{Acknowledgments}
This work was supported by 
National Key Research and Development Program of China (Grant No. 2020AAA0108004),
National Natural Science Foundation of China (Grant No. 61672440), 
Natural Science Foundation of Tianjin (Grant No. 19JCZDJC31400),
and
Natural Science Foundation of Fujian Province of China (Grant No. 2020J06001).

\bibliographystyle{model2-names}
\bibliography{PSSAttention}

\end{document}